%% file: root.tex
\documentclass[letterpaper, 10 pt, conference]{ieeeconf}  

\IEEEoverridecommandlockouts                              

\usepackage{graphicx}
\usepackage{epsfig} 
\usepackage{times} 
\usepackage{amsmath} 
\usepackage{amssymb}  
\usepackage{cite}
\usepackage{multirow}
\usepackage{multicol}
\usepackage{color}
\usepackage{float}
\usepackage{graphicx,subcaption,lipsum}

\DeclareMathOperator*{\argmin}{arg\,min}

\DeclareMathOperator*{\SE3}{\mathnormal{SE}(3)}
\DeclareMathOperator*{\SO3}{\mathnormal{SO}(3)}

\title{\LARGE \bf
GSLoc: Visual Localization with 3D Gaussian Splatting
}

\author{Kazii Botashev$^{1}$ $\:$ $\:$  Vladislav Pyatov$^{1}$ $\:$ $\:$   Gonzalo Ferrer$^{1}$ $\:$ $\:$ Stamatios Lefkimmiatis$^{2}$  
\thanks{$^{1}$The authors are with the Skolkovo Institute of Science and Technology (Skoltech), Center for AI Technology.
       {\tt\small \{kazii.botashev, vladislav.pyatov, g.ferrer\}@skoltech.ru}}%
\thanks{$^{2}$Stamatios Lefkimmiatis is with MTS AI, Russia. {\tt\small s.lefkimmiatis@mts.ai}}%
}

\begin{document}
\maketitle
\thispagestyle{empty}
\pagestyle{empty}

\input{0_abstract}

\input{1_introduction}

\input{2_related_work}

\input{3_background}

\input{4_method}

\input{5_results}

\input{6_ablation}

\input{7_limiatations_conclusions}

\section*{APPENDIX}
\section{IMPLEMENTATION AND RUNTIME DETAILS}
We modify the original CUDA-based implementation  \cite{kerbl20233d} of the differentiable renderer enabling camera pose-related gradients. We solve optimization using Adam \cite{adam} optimizer for 2000 steps or until convergence when the loss change is smaller than $10^{-5}$ for 3 consecutive iterations. On average, for cases that achieve a successful outcome the number of necessary iterations may vary between 100-300 iterations and take around 5-15 seconds to converge on a modern GPU. The optimization learning rate starts with $10^{-2}$ and exponentially decays with iterations to $10^{-5}$. The Gaussian blur is applied for the first 1000 iterations, its kernel covariance $\delta_j$ decays linearly from $10^{-1}$ to $10^{-4}$, its kernel size $L$ is 200 pixels for standard resolution experiments and is decreased for resolution ablation according to the image downscale factors. For the coarse-to-fine optimization strategy we decide that the localization successfully converged and should not be restarted if the rendered image PSNR has reached 25 dBs.

\input{8_acknowledgements}

\addtolength{\textheight}{-12cm}   

\end{document}

%% file: 0_abstract.tex
\begin{abstract}
We present GSLoc: a new visual localization method that performs dense camera alignment using 3D Gaussian Splatting as a map representation of the scene. 
GSLoc backpropagates pose gradients over the rendering pipeline to align the rendered and target images, while it adopts a coarse-to-fine strategy by utilizing blurring kernels to mitigate the non-convexity of the problem and improve the convergence. The results show that our approach succeeds at visual localization in challenging conditions of relatively small overlap between initial and target frames inside textureless environments when state-of-the-art neural sparse methods provide inferior results. Using the byproduct of realistic rendering from the 3DGS map representation, we show how to enhance localization results by mixing a set of observed and virtual reference keyframes when solving the image retrieval problem. We evaluate our method both on synthetic and real-world data, discussing its advantages and application potential.


\end{abstract}

%% file: 1_introduction.tex
\section{INTRODUCTION}

Visual localization, the process of determining the camera pose using a visual representation of a known scene, plays an important role in various applications related to robot navigation, self-driving cars, and augmented/virtual reality \cite{lynen2015get, heng2019project}. In particular, the main objective of visual localization via camera alignment is, provided an input query image, to determine the 6 degrees of freedom (dof) camera pose (position and orientation) in a 3D environment with a known map representation.

The map representation of a known scene, which is a core part of every localization method, can be of different forms. The most developed and commonly used ones are sparse map representations \cite{collet2009object, li2018deepim}, which rely on a set of 2D-3D feature-landmark correspondences typically estimated using structure-from-motion (SfM) techniques \cite{schonberger2016structure}. Despite their effectiveness in various localization scenarios, sparse map representations provide limited scene comprehension, falling short in empty spaces or textureless environments with no distinct features. Dense mapping is an alternative family of representations that aim to utilize information from entire images but may require capturing depth, ensuring continuity of the input frames \cite{engel2014lsd, engel2017direct, forster2014svo, zubizarreta2020direct}. Other methods may operate on dense image descriptors \cite{taira2018inloc, Arandjelovic16}, usually extracted with convolutional neural networks (CNN). Methods of this category have proven their efficiency in large-scale scenarios and image retrieval tasks but have limited accuracy and produce only an approximated pose of the query camera.

Differentiable mesh-based rendering algorithms have also been employed for the visual localization task, leading to a family of dense map representation methods that can achieve impressive results. However, this comes at the significant cost of requiring a detailed 3D model of the environment \cite{chen2020category, park2020latentfusion}. This drawback has been recently mitigated with the introduction of Neural Radiance Field (NeRF) \cite{mildenhall2021nerf} models that can be trained using only a set of posed images. NeRFs can achieve photo-realistic rendering quality by implicitly learning via 2D supervision the 3D scene as a function of a continuous radiance field.  While NeRFs were originally introduced to deal with novel-view synthesis, their learned map representation has been recently used in the design of novel pose estimation methods. Started with a simple idea presented in iNeRF \cite{yen2021inerf}, it continued with other sophisticated pose estimation approaches \cite{maggio2023loc, sucar2021imap}. However, despite their initial promising results, such methods still face a limited applicability since they suffer from the same drawbacks of NeRF models, that is extremely long training and rendering times due to the expensive backward mapping ray-casting procedure.

Recently, 3D Gaussian Splatting (3DGS) \cite{kerbl20233d} has been introduced and achieves high-quality real-time novel view synthesis at full HD resolution. This is an alternative learning-based approach that unlike NERF-based methods is based on a forward mapping/rasterization strategy. Specifically, 3DGS represents the 3D scene with a collection of 3D anisotropic Gaussians, which play the role of  rendering primitives and whose parameters are directly optimized from a set of available posed images during training. The type of operations required by a 3DGS rasterizer are better suited for GPUs resulting in a very efficient and interactive novel view rendering process.

3DGS introduces a novel and distinctive map representation of the environment, which shows promise for effectively addressing the challenges associated with camera pose estimation and visual localization. The 3DGS strategy is computationally efficient and fully differentiable, facilitating the generation of highly realistic images in arbitrary views. Importantly, it allows for the direct flow of parameter gradients for any given camera pose, enabling real-time dense camera alignment, a capability not offered by other localization methods. Furthermore, it establishes a unique and fully-differentiable rendering-pose relation, enabling the generation of rendering images for any given camera and facilitating gradient-based optimization to refine its pose by minimizing the discrepancy between rendered and query images.

Nevertheless, there are still two challenges associated with this novel approach. The first one is that the accuracy of the initial camera pose used during training can have a significant impact on the success of the method. Secondly, the utilized objective loss, which is based on the photometric difference, is highly non-convex due to the presence of high-frequency details in the images. The non-convex nature of the loss poses difficulties in its optimization, as it can lead to the entrapment of the optimization process to one of the numerous local minima, resulting in suboptimal solutions.

This work focuses on utilizing the 3D Gaussian Splatting rendering technique as a map representation for visual localization tasks and aims to overcome the existing challenges described above. Our study includes an investigation of the viability of the 3DGS method as a map representation, a comprehensive convergence analysis for various camera initialization scenarios, an exploration of convergence limitations arising from the highly non-convex nature of the problem, and the proposal of a coarse-to-fine optimization strategy to mitigate such limitations. The main contributions of this work are summarized as follows:

\begin{itemize} 
\item We analytically derive the gradients of the 3DGS renderings with respect to to camera poses and implement a 3DGS-based visual localization pipeline.

\item We propose a coarse-to-fine optimization strategy where we apply gradually fading Gaussian blur on the query and rendered images that allows us to overcome the problem of suboptimal convergence for high-frequency image details.   

\item We propose an effective way of improving the localization results by enhancing camera initialization obtained via image retrieval by extending its image base with rendered camera frames.

\item We evaluate our approach on indoor synthetic and real scenes, provide a comprehensive quantitative analysis of camera pose convergence based on various initial camera pose priors and parameterizations, and compare it with a sparse feature-based localization baseline.
\end{itemize}

%% file: 2_related_work.tex
\section{RELATED WORK}

\subsection{Visual Localization methods}
Classic sparse feature-based localization focuses on detecting and matching a sparse set of distinctive features or keypoints in the camera images \cite{collet2009object, li2018deepim}. The initial approach to feature matching involved manual design of keypoint detection algorithms that can identify visually salient image details: points, edges, and corners \cite{bay2008speeded}. However, the recent progress in the field, which has been driven by the introduction of dedicated neural networks for feature extraction \cite{sarlin2020superglue}, has led to a revision of the feature extraction and matching stages. Indeed, these network architectures have demonstrated exceptional performance, achieving precise and robust feature matching results.

Consequently, the map representation in sparse feature-based localization can be constructed using network-extracted keypoints alongside their related 3D positions or descriptors. At the localization stage, the query image is processed to extract keypoints that are afterwards matched against the map to estimate the 6-DoF camera pose. Sparse feature-based methods are computationally efficient and have demonstrated robustness in a variety of applications. However, they cannot be used for tasks that require scene understanding while they also disregard useful volumetric context and, thus, can fail in featureless or empty environments.

On the other hand, dense visual localization methods aim to utilize an entire image dense map representation by matching visual information across the entire images using dense descriptors, such as pixel-level descriptors or dense feature maps. This type of representations encodes information about the appearance, texture, or semantic context of the scene \cite{Arandjelovic16}.

\subsection{Rendering-based Pose Estimation}

The introduction of Neural Radiance Fields (NeRF)~\cite{mildenhall2021nerf} and the large array of follow-up work\cite{yen2021inerf, maggio2023loc, sucar2021imap} has brought a new paradigm to novel view synthesis and subsequently to camera pose estimation. In particular, NeRF represents the scene as a continuous 3D volume and learns the radiance field properties, resulting in a more accurate and realistic map representation of static scenes with intricate geometry and complex lighting. 
While NeRF-based camera pose estimation methods have certain advantages and can achieve competitive results, they also face a limited applicability due to the principal drawbacks of the NeRF model itself, including long model training and inference time due to the computationally expensive utilized backward mapping/ray-casting procedure.

Meanwhile, recent studies indicate that 3DGS-based camera pose estimation methods effectively circumvent these drawbacks and hold significant promise for future applications \cite{gsslam, sun2023icomma}.

%% file: 3_background.tex
\section{RENDERING WITH 3D GAUSSIAN SPLATTING}

For a given set of $\mathcal{N}$ RGB images $\{I_k\}_{k=1}^{\mathcal{N}}$ and corresponding camera poses ${T_{w}^{c}}_k\in\SE3 $, 3DGS can learn a 3D scene representation that enables photo-realistic novel view rendering for an arbitrary camera pose. This is achieved by modeling the scene using a collection of $\mathcal{M}$ 3D Gaussians, which are defined in a world coordinate frame $w$:
\begin{equation}
\mathbf{G} = \{ G_i^w : ( \boldsymbol{\mu}_i^w, \boldsymbol{\Sigma}_i^w, 
\boldsymbol{\sigma}_i, 
\boldsymbol{c}_i
)  \}_{i=1}^{\mathcal{M}}.
\label{gaussians}
\end{equation}
These Gaussians serve as rendering primitives and are fully described by their centers $\boldsymbol{\mu}_i^w \in \mathbb{R}^3$, covariance $\boldsymbol{\Sigma}_i^w \in \mathbb{R}^{3 \times 3}$, opacity $\boldsymbol{\sigma}_i \in \mathbb{R}$ and view-dependent color $\boldsymbol{c}_i \in \mathbb{R}^3$.

Having the 3D Gaussian scene representation $\mathbf{G}$ at hand, rendering the image for a novel view, which is determined by a camera pose defined using a world $w$ to camera $c$ rigid body transformation $T_{w}^{c}= \{ \mathbf{R}_w^c \in \SO3, \mathbf{t}_w^c \in  \mathbb{R}^3 \} \in \SE3 $, proceeds by perspectively projecting the Gaussians $G_i^w$ to the image plane $\mathcal{I}$ in the ray space. To do so, we first express the Gaussians w.r.t the camera frame, which leads to:
\begin{equation}
\boldsymbol{\mu}_i^c = T_{w}^{c} \boldsymbol{\mu}_i^w ; \:  \: \: \boldsymbol{\Sigma}_i^c = \mathbf{R}_w^c \boldsymbol{\Sigma}_i^w {\mathbf{R}_w^c}^\intercal
\label{world_to_cam}
\end{equation}
Then, all the Gaussians are perspectively projected to the image plane in the ray space using an affine approximation of the projective non-linear transformation $\pi$ that involves its Jacobian $\mathbf{J}$. This approximation leads to a mapping of the initial 3D Gaussians to 2D Gaussians whose centers and covariances are expressed as:
\begin{equation}
\boldsymbol{\mu}_i^{\mathcal{I}} = \pi( \boldsymbol{\mu}_i^c) ; \: \: \: \boldsymbol{\Sigma}_i^{\mathcal{I}} =  \mathbf{J} \boldsymbol{\Sigma}_i^c \mathbf{J}^\intercal = 
\mathbf{J} \mathbf{R}_w^c \boldsymbol{\Sigma}_i^w {\mathbf{R}_w^c}^\intercal \mathbf{J}^\intercal 
\label{cam_to_image}
\end{equation}
Finally, the image intensity $\hat{\mathbf{C}}$ is computed via depth-ordered $\alpha$-blending of the projected Gaussians as follows:
\begin{equation}
\hat{\mathbf{C}}=\sum_{i \in \mathcal{M}} \mathbf{c}_i(\mathbf{d}_i) \alpha_i \prod_{j=1}^{i-1}\left(1-\alpha_j\right),
\label{alpha_blending}
\end{equation}
where the density $\alpha_i$ is computed as a multiplication of the covariance $\boldsymbol{\Sigma}_i^{\mathcal{I}}$ and opacity $\boldsymbol{\sigma}_i$,
$\mathbf{c}_i(\mathbf{d}_i)$ is the view-dependent color of the 3D Gaussians defined with spherical harmonics and computed based on the view-direction vector $\mathbf{d}_i =  (\boldsymbol{\mu}_i^w - \mathbf{t}_c^w) / \|(\boldsymbol{\mu}_i^w - \mathbf{t}_c^w) \| $.

Starting with some sparse SfM \cite{schonberger2016structure} point cloud initialization and following the above described fully-differentiable rendering procedure, 3DGS gradually optimizes the 3D Gaussian parameters with gradient descent by minimizing the weighted combination of $L_1$ and D-SSIM losses between the rendered image $\hat{I}_k({T_{w}^{c}}_k, \mathbf{G})$ and the ground-truth posed image ${I}_k$. As a result, the 3D scene is learned with a high fidelity representation allowing for photo-realistic novel view rendering.

%% file: 4_method.tex
\section{METHOD}
The learned 3DGS scene model $\mathbf{G}$ serves as a novel and distinctive map representation of the 3D environment and is potentially highly suitable for being utilized in the visual localization task. Specifically, for a query image $\Tilde{I}$ the corresponding pose $\Tilde{T}_w^c = \Tilde{T} \in \SE3$ can be found by minimizing the discrepancy between the rendered $\hat{\Tilde{I}}$ and query images:
\begin{equation}
\Tilde{T}=\underset{T \in \SE3} \argmin \,{  \mathcal{L}_1(\hat{\Tilde{I}}(T, \mathbf{G}), \Tilde{I})}.
\label{argmin_pose_estimation}
\end{equation}
Ostensibly, the solution of this task seems to be straightforward thanks to the photo-realistic real-time rendering capabilities of 3DGS. However, there are several aspects that require specific attention and which we address next. 

\subsection{Camera Pose Gradients}
To achieve real-time rendering performance, 3DGS utilizes GPU capabilities and its official implementation of the rasterization step is based in CUDA. This precludes out-of-the-box automatic differentiation and instead requires the derivation of the explicit form of gradients for all the parameters to be optimized. To enable camera pose optimization, it is also required to express analytically all the gradients related to the poses parameters. 

Since in the original work of 3DGS the authors did not optimize the camera poses we need to derive all the gradients related to the pose parameters that we wish to estimate. In this work, we
implement the camera pose optimization on the Riemannian manifold and use Lie algebra to derive the camera pose Jacobians for the terms in Eq. \eqref{cam_to_image} and Eq. \eqref{alpha_blending} via the chain rule as follows:
\begin{equation}
\frac{\partial \boldsymbol{\mu}_i^{\mathcal{I}}}{\partial T_{w}^{c}}=\frac{\partial \boldsymbol{\mu}_i^{\mathcal{I}}}{\partial \boldsymbol{\mu}_i^c} 
\frac{\partial \boldsymbol{\mu}_i^c}{\partial T_{w}^{c}}
\label{chain_rule_mean}
\end{equation}
\begin{equation}
\frac{\partial \boldsymbol{\Sigma}_i^{\mathcal{I}}}{\partial T_{w}^{c}}=\frac{\partial \boldsymbol{\Sigma}_i^{\mathcal{I}}}{\partial \mathbf{J}} \frac{\partial \mathbf{J}}{\partial \boldsymbol{\mu}_i^c}\frac{\partial \boldsymbol{\mu}_i^c}{\partial T_{w}^{c}}+\frac{\partial \boldsymbol{\Sigma}_i^{\mathcal{I}}}{\partial \mathbf{R}_w^c} \frac{\partial \mathbf{R}_w^c}{\partial T_{w}^{c}}
\label{chain_rule_cov}
\end{equation}
\begin{equation}
\frac{\partial \mathbf{c}_i}{\partial T_{w}^{c}}=\frac{\partial \mathbf{c}_i}{\partial \mathbf{d}_i} \frac{\partial \mathbf{d}_i}{\partial \mathbf{t}_c^w}
\frac{\partial \mathbf{t}_c^w}{\partial T_{w}^{c}}.
\label{chain_rule_color}
\end{equation}
We compute the derivatives on the manifold following the general approach detailed in \cite{botashev2023analytical}. Due to space limitations we omit their detailed derivations and provide only their final forms:
\begin{equation}
\frac{\partial \boldsymbol{\mu}_i^c}{\partial T_{w}^{c}} = \left[\begin{array}{ll}
\mathbf{I} & -{\boldsymbol{\mu}_i^c}^{\wedge}
\end{array}\right] ; \: \: \frac{\partial \mathbf{t}_c^w}{\partial T_{w}^{c}} = \left[\begin{array}{ll}
{\mathbf{R}_w^c}^\intercal  & \mathbf{0}
\end{array}\right]
\label{gradient_mean}
\end{equation}
\begin{equation}
\frac{\partial \mathbf{R}_w^c}{\partial T_{w}^{c}}=\left[\begin{array}{cc}
\mathbf{0} & -\mathbf{r}_{c1}^{\wedge} \\
\mathbf{0} & -\mathbf{r}_{c2}^{\wedge} \\
\mathbf{0} & -\mathbf{r}_{c3}^{\wedge}
\end{array}\right]
\label{gradient_cov}
\end{equation}
where $^\wedge$ denotes the skew symmetric matrix constructed from the corresponding input vector and $\mathbf{r}_{cj}$ denotes the j-th column of the rotation matrix $\mathbf{R}_w^c$.

Using the equations \eqref{chain_rule_mean}-\eqref{gradient_cov} it is possible to propagate all the necessary gradients for the camera pose optimization task. We iteratively solve the optimization problem of \eqref{argmin_pose_estimation} for 2000 steps or until convergence using the first-order Adam \cite{adam} optimizer with exponentially decaying learning rate. More details are provided in the appendix.

Manifold-derived pose Jacobians have a minimal 6 DoF representation and lead to a better convergence compared to alternative parametrizations. In particular, according to our ablation study, the manifold optimization achieves better results and shows clear advantages compared to the common alternative parametrization utilizing quaternions for rotation and 3D vectors for translation.

\subsection{Impact of Initial Camera Pose Proximity}
\begin{figure}[thpb]
 \centering
\includegraphics[width=0.9\linewidth]{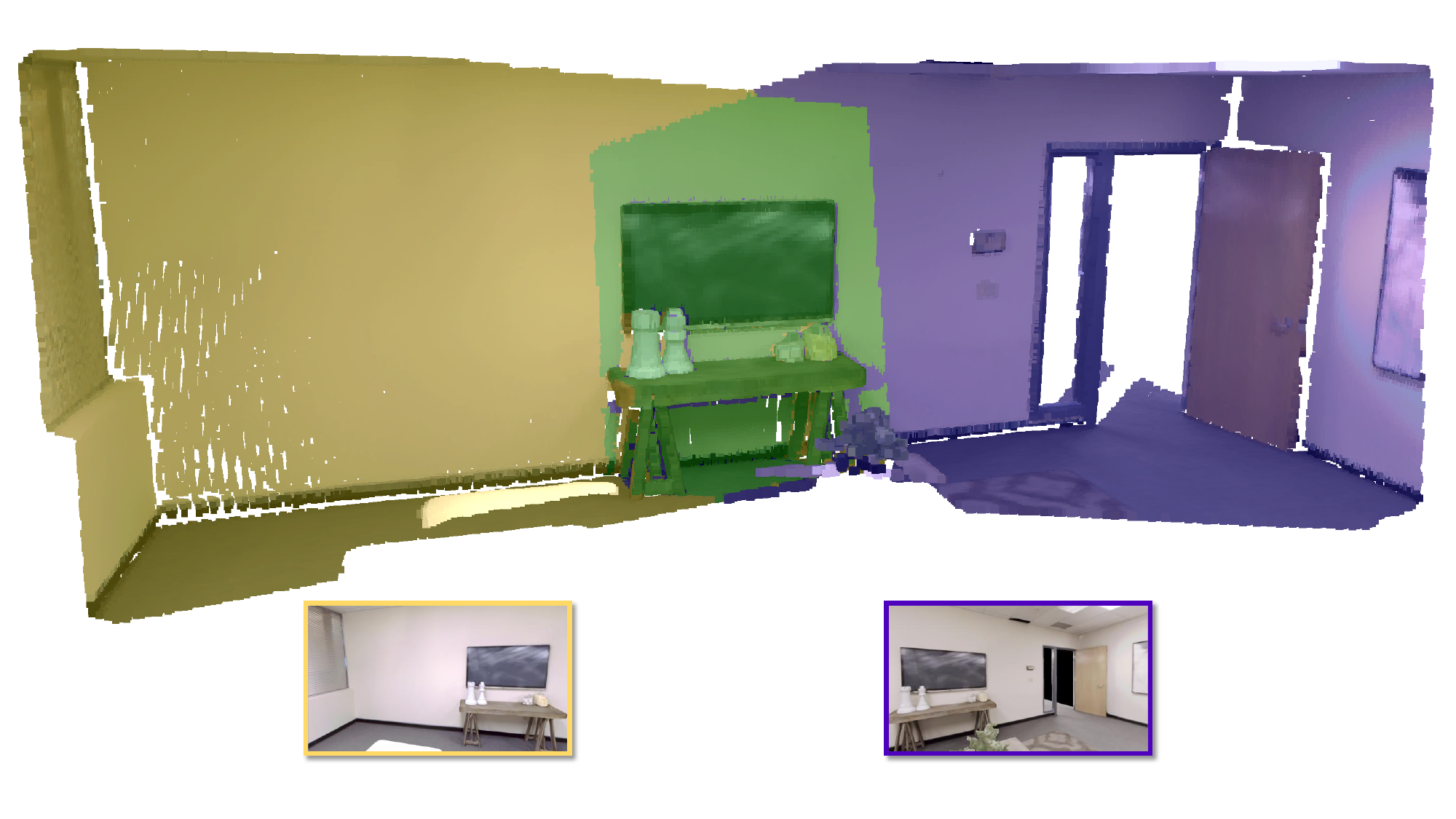}
  \vspace{-2mm}
  \caption{Visual explanation of 3D Intersection over Union (IoU) metric used for camera frames proximity estimation. Computed with voxels of the scene, this metric naturally describes both proximity of the camera poses and the visual similarity of their image frames. Here for the visualized frames the 3D IoU is equal to 0.15.}
  \label{fig:ious}
    \vspace{-3mm}
\end{figure}

Among the most important factors that affect the final result of visual localization is the proper initialization of the camera poses. Finding an initial camera pose that exhibits a sufficiently large overlap with the query camera frame is crucial and can be one of the main factors of success or failure. This problem, which is also known as the Image Retrieval task, is a separate long-standing computer vision problem that requires special attention on its own. While, a 3DGS-based solution of this task is an intriguing possible research direction, here we focus on solving the exact visual localization task. As a result, finding the best possible existing Image Retrieval algorithm for the camera pose initialization for GSLoc is out of scope of this work.  

Instead, we seek to perform a comprehensive analysis of our method. We aim to estimate the dependency between obtaining the correct result with GSLoc and the proximity of the initial camera frame to the target one. In other words, we want to answer the following questions: 1) How close/far our initial guess of the camera pose need to be so that our method leads to the correct solution? and 2) What are the chances of this convergence?

\begin{figure*}[thpb]
 \centering
 \vspace{1.5mm}
\includegraphics[width=.99\textwidth]{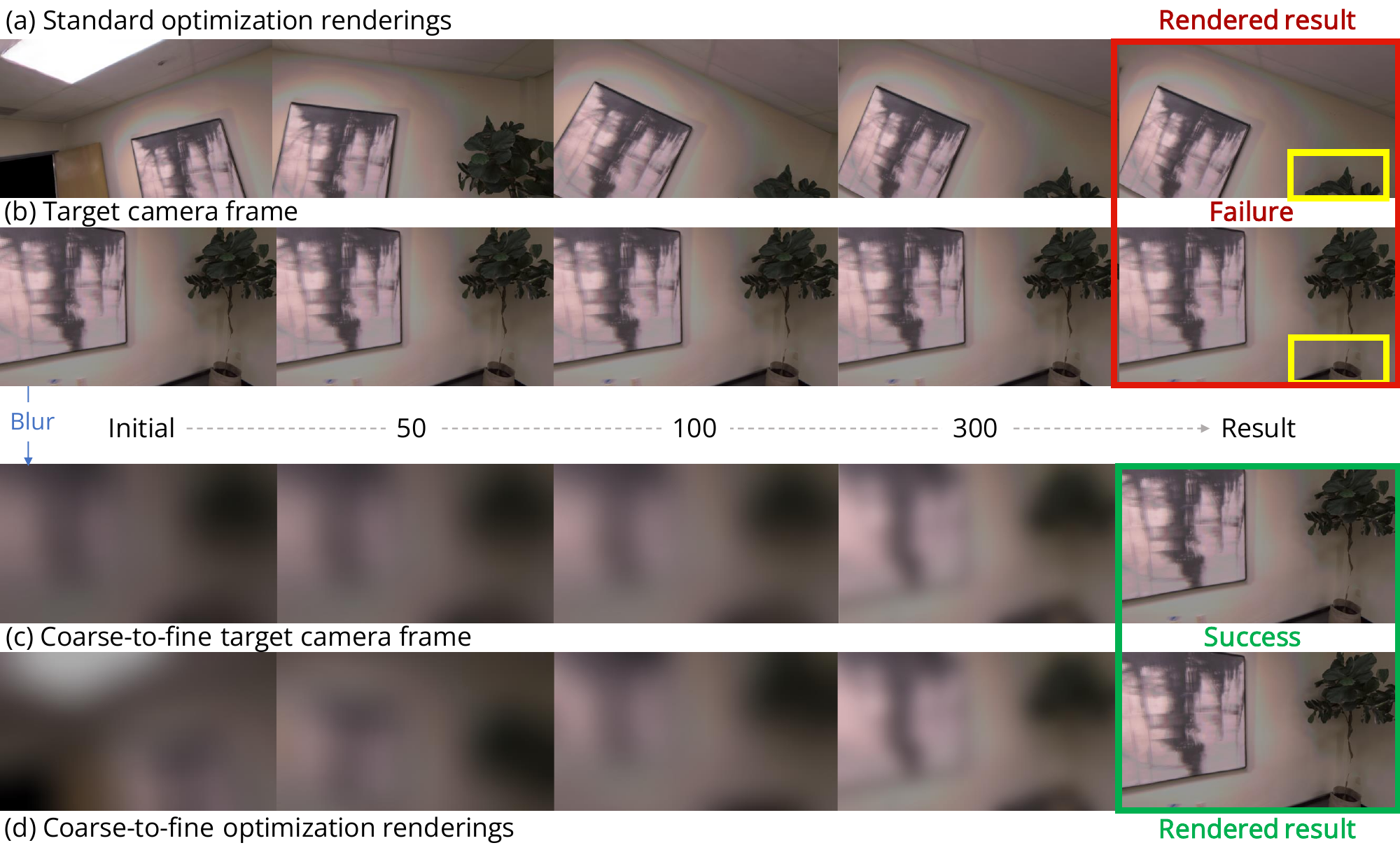}
\vspace{-1.5mm}
  \caption{Visualization of the camera pose alignment process induced by iterative optimization of photometric loss between intermediate renderings and target images for standard (a)-(b) and coarse-to-fine (c)-(d) strategies. Standard optimization decsribed with (a)-(b) leads to convergence to a sub-optimal solution: it does not manage to escape the local minima caused by the sub-optimal overlap between the intermediate rendering and the target query image (highlighted with yellow) resulting to an unsuccessful image alignment. On the contrary, smoothing the image gradients with our coarse-to-fine approach (c)-(d) allows us to avoid being trapped in local minima and converge to the correct camera pose.}
  \label{fig:coarse_fine}
  \vspace{-5mm}
\end{figure*}

To answer these questions, we propose to measure the camera frames proximity using the 3D Intersection over Union (IoU) metric that is computed using voxels of the scene. As visualized in Fig.~\ref{fig:coarse_fine}, this metric naturally describes both proximity of the camera poses and the visual similarity of their corresponding image frames and enable us to quantitatively assess their impact on the final result. For our particular task the 3D IoU is more informative and intuitive compared to simple rotation and translation distances. 

\subsection{Extending image retrieval database with renderings}
Besides the 3D IoU criterion, we also evaluate the results of the visual localization for initial poses corresponding to the closest dense image descriptors. Following a widely adopted approach, we extract  global image descriptors using NetVLAD \cite{Arandjelovic16} and for each query image we find the most similar ones in the pool of images used for 3DGS training. Next, we solve the visual localization task by initializing the camera pose with these closest matches. 

This is a very common approach widely adopted as a first step for sparse feature based visual localization. Its results directly depend on the number and diversity of images used as a map for comparison. However, 3DGS-based map representation allows us to overcome this limitation by extending the original set of images used both for 3DGS training and image retrieval step with any number of posed photorealistic scene renderings produced with the optimized 3DGS scene. Starting with a limited set of images, 3DGS allows us to extend our image base used for Image Retrieval by creating arbitrary novel-view renderings. This increases the probability of obtaining a good initial pose and as a result increases our chances of converging to the correct solution. Further, we show the effectiveness of this proposed technique both for synthetic and real scenes in evaluation.

\subsection{Coarse-to-fine Rendering Scheduling}
The highly non-convex nature of the photometric $\mathcal{L}_1$ loss w.r.t the 6 DoF space of camera poses in $\SE3$ poses a significant challenge related to its optimization. This non-convexity is caused by high-frequency image details and can lead to the entrapment of the optimization process to a bad local minima, which in turn can lead to a sub-optimal solution.

The visual representation of one such case is depicted in Fig.~\ref{fig:coarse_fine}(a)-(b). Iteratively minimizing the  objective function in \eqref{argmin_pose_estimation} in a standard way leads to the convergence of the first-order method to a sub-optimal solution. Indeed, it is clearly visible that during standard optimization using Adam \cite{adam} does not manage to escape the local minima caused by the sub-optimal overlap between the intermediate rendering and the target query image. This results to an unsuccessful image alignment (highlighted with yellow). 

To overcome this problem we propose a simple yet effective coarse-to-fine strategy of applying a progressively decaying Gaussian blur both on the rendered and target images. Specifically, we convolve both the target query image and the intermediate rendering with a 2D Gaussian kernel 
$\mathcal{N}_{2d}(\delta_j)\in\mathbb{R}^{L \times L} $ of fixed size $L$ while gradually decreasing its covariance $\delta_j$. This results in a modified objective function $\mathcal{L}_1(\mathcal{N}_{2d} \ast\hat{\Tilde{I}}(T, \mathbf{G}), \mathcal{N}_{2d} \ast \Tilde{I})$ with smoothed gradients and a stabilized camera pose estimation as depicted in Fig.~\ref{fig:coarse_fine}(c)-(d). Smoothing the image gradients allows us to avoid being trapped in local minima and converge to the correct camera pose. 
We have also found it to be effective the strategy of running several passes of coarse-to-fine optimization, restarting each new pass with the result from the previous one.
Based on the above, we have concluded that the highest efficiency is achieved by the following two-step GSLoc algorithm: 1) In the first step a standard camera pose optimization takes place. 2) If the first step does not recover the correct pose (the photometric loss between the rendered image and the query image exceeds a user-defined threshold), then we restart the entire process and apply the described coarse-to-fine optimization strategy.

%% file: 5_results.tex
\section{EVALUATION}
We assess the performance of our proposed method by performing extensive experiments on 5 synthetic scenes from the Replica \cite{straub2019replica} dataset. Our motivation for using this data in our evaluation stems from several reasons that we discuss next. The first reason is that the use of synthetic data ensures that the  3DGS map representation $\mathbf{G}$ is adequately learned. This can be achieved by exploiting the available accurate ground truth poses and depth information during the 3DGS training. In turn, this allows us to neglect any possible negative effects of the incorrect map representation on the visual localization results and validate the efficiency of the proposed method without worrying about data-related inaccuracies. Another reason is that we have access to the ground truth poses which allows us to accurately compute the localization errors and perform a precise evaluation of the proposed method. Finally, in order to conduct a comprehensive study of the effect of pose initialization based on the camera proximity according to the 3D voxel-based IoU, we need access to the detailed 3D voxel model of the scene, which we can accurately extract from such synthetic data. Nevertheless, we also evaluate our method on 2 real scenes from the Deep Blending dataset \cite{hedman2018deep} and show the coherence of the obtained real-data results with the synthetic ones.

\subsection{Synthetic Data and Initial Camera Analysis}
\subsubsection{Setting}
We perform our study using 5 scenes from the Replica \cite{straub2019replica} dataset. For each scene we assume that the camera intrinsics are known and utilize a base trajectory of approximately 200 frames that describe in detail the environment. These frames are accompanied by a depth estimated point cloud and camera poses. 

Next, for each scene we capture a diverse set of 32 test query frames that we then use for the visual localization evaluation. For each of these query frames we randomly generate 16 different pose initializations related to each one of 10 different 3D IoU levels lying in the range between 0.05 and 0.65. Overall, we utilize 5 scenes - 32 query frames - 10 IoU levels - 16 different initalization poses. In total this amounts to approximately 25k different localization tasks, which allow us to thoroughly validate the performance of GSLoc, its strengths and possible limitations. 

\begin{figure*}[ht]
 \centering
 \vspace{2.25mm}
  \subfloat[Rotation results]{\includegraphics[width=.42\textwidth]{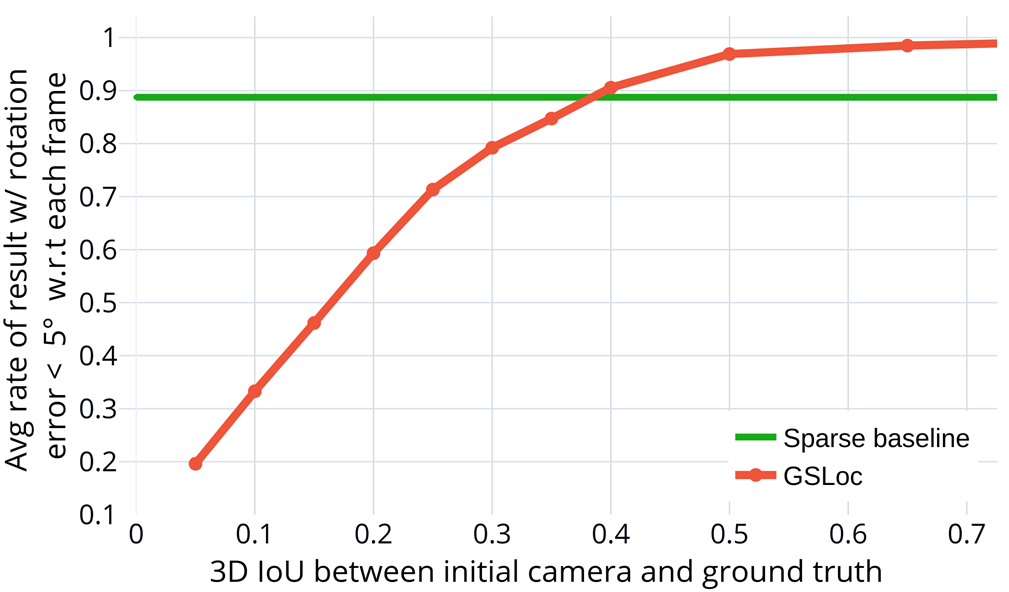}
  \hspace{.05\textwidth}
  }
  \subfloat[Translation results]
  {\includegraphics[width=.42\textwidth]{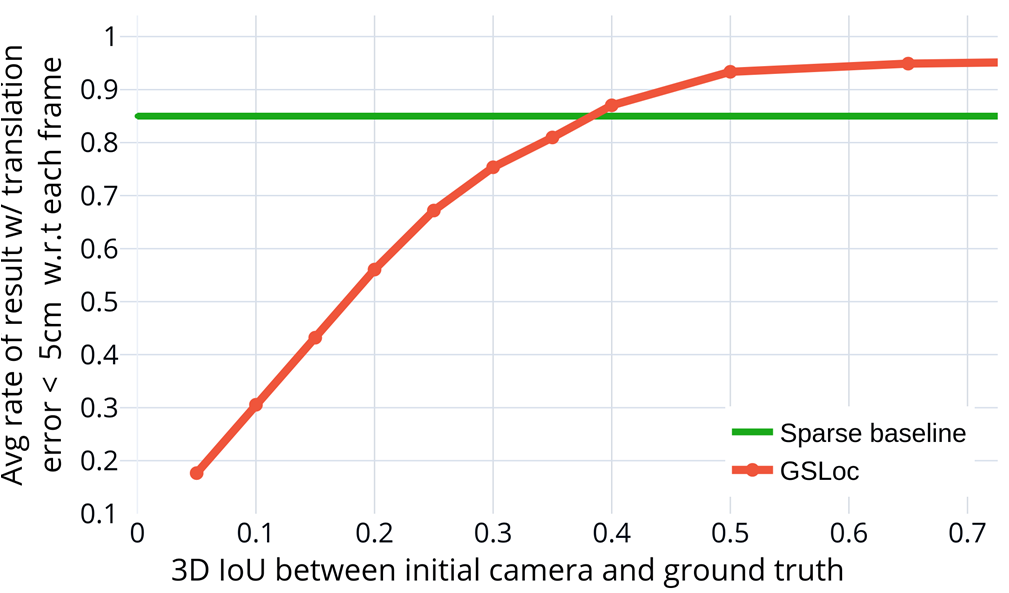}}
  \vspace{-1.5mm}
  \caption{Quantitative results of GSLoc on synthetic scenes from Replica \cite{straub2019replica} dataset compared with sparse feature-matching baseline. Provided results show the dependency between obtaining the correct pose with GSLoc and the proximity of the initial camera frame to the target one. With the increase of the frames' proximity, GSLoc first reaches and then surpasses the baseline. We report the results separately for rotation (a) and translation (b) pose components.}
  \vspace{-5mm}
\label{fig:main_result}
\end{figure*}

In order to not significantly deviate from a realistic setup, we process the base images using SfM to estimate the camera poses and the 3D points, with hloc serving as the SfM reconstruction method. Specifically, we first extract local feature descriptors with SuperPoint \cite{detone2018superpoint} and global image descriptors with NetVLAD \cite{Arandjelovic16}. Based on the similarity of NetVLAD decriptors we then estimate the top 5 neighbors for each image. After that we use SuperGlue to perform feature matching of each image with it’s top 5 neighbors. The rest of the reconstruction is performed with COLMAP’s \cite{schonberger2016structure} incremental mapping.

This SfM reconstruction is a priori not perfect due to the inherent flaws of such methods. Therefore, in order to minimize the impact of these map reconstruction errors on visual localization, we replace all the poses and 3D points in the SfM reconstruction with the ground truth ones. One can interpret this choice as if we had a flawless SfM reconstruction. With this strategy, we manage to avoid overly refined experiments setup, keeping the evaluation clean but still realistic. After this, we use this clean SfM reconstruction both for learning the 3DGS scene map representation and as a database for the sparse matching based visual localization baseline. The localization baseline that we use for comparison consists of the following steps. For the query image $I_q$ we perform feature extraction with SuperPoint \cite{detone2018superpoint}. Then we extract its NetVLAD \cite{Arandjelovic16} global descriptor, find the top 5 neighbors from the database and perform feature matching with SuperGlue. This gives us 2D-3D correspondences that we use in PnP RANSAC to estimate the absolute pose. The final pose of the query image is obtained after the non-linear refinement with the Levenberg-Marquardt algorithm.

\subsubsection{Results}
For all the 5 scenes, we have in total 160 test query frames to localize. Following \cite{yen2021inerf}, we identify the localization as successful if the resulting pose error is less than 5 degrees for rotation and 5cm for translation. While for the baseline the result is binary: either success or failure, for GSLoc method we separately evaluate each of 10 IoU proximity levels trying to localize each frame with 16 different initializations. Hence, for each IoU level the result of localizing each of the 160 test frames is not binary but the 0-1 ratio describing the percentage of successful results for 16 different initialization per query frame. By averaging this ratios w.r.t to all query frames, we evaluate the efficiency of our method and present the results along with the baseline comparison in Fig.~\ref{fig:main_result}.

These results indicate that with an increased proximity of the initial camera pose the 3DGS based visual localization improves its results getting close and outperforming the sparse feature based localization after reaching the threshold of $\sim{0.4}$ 3D IoU. Although, it may seem that the baseline results are not perfect, it has in fact managed to solve most of the localization tasks, failing only for extreme featureless images. While such cases are almost impossible to solve with sparse methods, GSLoc handles them well due to its dense alignment nature.
Based on the above, we conclude that the 3DGS based map representation used by GSLoc is suitable for solving visual localization tasks and can lead to competitive results. 
We also show that as any other visual localization methods, GSLoc requires a certain level of proximity of the initial camera pose used for optimization. We elaborate on this aspect of the problem in the next section.

\subsection{Enhanced Camera Initialization with Image Retrival on Extended Image Base}
 The camera initialization for visual localization problems is typically obtained by solving the Image Retrieval task. A common way to do this is by finding the closest image descriptors in an image database. For instance, such approach corresponds to the preliminary step used in our sparse baseline. In this section we investigate whether such resulting poses are suitable enough for being used as initialization within GSLoc.

\subsubsection{Setting}
We follow the same setup as the one used in the previous experiment. The only major difference here is that instead of investigating different 3D IoU levels for initial camera poses, we simply follow our sparse baseline and initialize the pose with those that have 5 closest NetVlad \cite{Arandjelovic16} descriptors among images in our base trajectory. Here we switch on the binary result classification, assuming the localization of the query to be successful if at least one out of five camera initialization lead GSLoc to the correct final solution.

Further, we utilize the learned 3DGS scenes to extend our base trajectory image databases used for the image retrieval task. We carefully capture an additional set of camera frames extending the existing database and increasing its diversity and scene comprehension.

\subsubsection{Results}
\begin{figure}[thpb]
 \centering
 \vspace{1.75mm}
\includegraphics[width=0.88\linewidth]{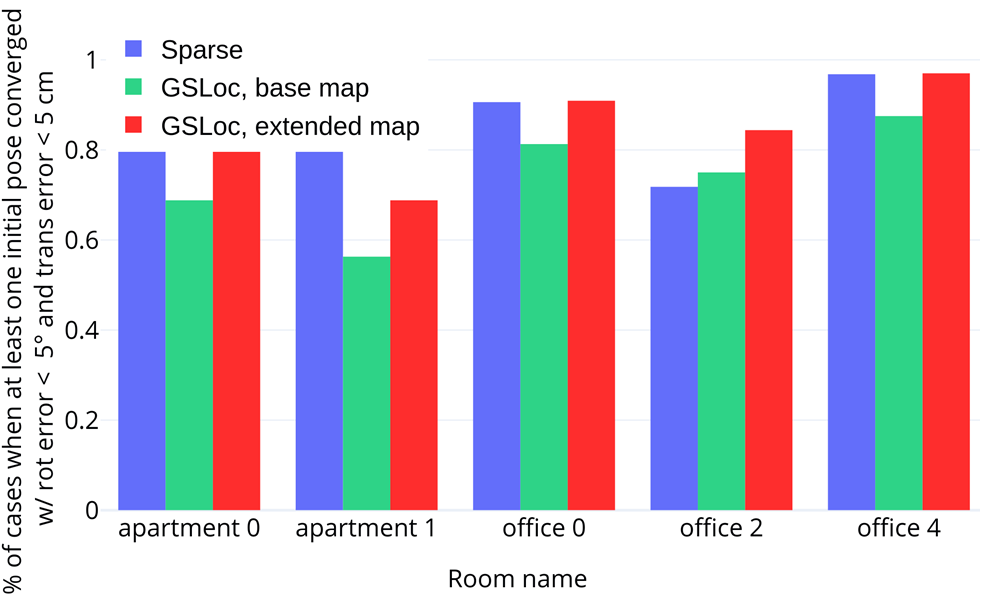}
\vspace{-1.5mm}
  \caption{Quantitive results on the synthetic scenes from Replica \cite{straub2019replica} dataset. Enhancing the GSLoc camera initializations obtained by the image retrieval with the rendering-extended imagebase leads to consistent success rate improvement proving the efficiency of the proposed method.}
  \vspace{-5mm}
  \label{fig:base_rebase}
\end{figure}

We report the results for the experimental setup described above in Fig.~\ref{fig:base_rebase}. We describe the percentage of the successfully localized frames for each individual room of Replica \cite{straub2019replica} dataset. Detailing the previously reported results, the baseline method initialized with closest descriptors manages to achieve the correct result for most of the cases, failing only for frames that are dominated by empty space as in the ``office 2" scene. 

In contrast, GSLoc initialized with the original base map dense matches, exhibits a slightly worse performance, on average having less successfully localized results. What is interesting is that extending the image base with the renderings actually helps to improve its performance and on average increases the GSLoc success rate by $10\%$ bringing it closer to the baseline. Investigating the reasons for this improvement, we figured out that the median 3D IoU for the camera initializations estimated with dense matching on the original database is $\sim0.3$ and increases to $\sim0.4$ with the database rendering extension. This is an indirect confirmation of the results previously reported in Fig.~\ref{fig:main_result}. This experiment proves the efficiency of the proposed base map extension technique and reveals a potential of utilizing 3DGS-based methods for successfully solving the image retrieval task.

Concluding the discussion of the results obtained on synthetic data, we report the average numerical pose estimation errors of successfully localized frames for all previously reported experiments in Tab.~\ref{tab:metrics}. We show that the successful localization with GSLoc leads to comparable or even smaller pose errors compared with the sparse baseline method. Besides that, we also show that successfully localized frames also achieve higher visual metrics, which directly relate to the quality of the pose estimation and therefore might be used as a criterion for deciding if localization has been successful. 

\begin{table}[thpb]
\resizebox{\linewidth}{!}{%
\begin{tabular}{|c|cc|cc|c|}
\hline
\multirow{2}{*}{Method} & \multicolumn{2}{c|}{Rotation error, deg} & \multicolumn{2}{c|}{Translation error, cm} & \multirow{2}{*}{\begin{tabular}[c]{@{}c@{}}Mean \\ PSNR, dB\end{tabular}} \\ \cline{2-5}
 & \multicolumn{1}{c|}{Mean} & Median & \multicolumn{1}{c|}{Mean} & Median &  \\ \hline
Sparse baseline & \multicolumn{1}{c|}{0.098} & 0.058 & \multicolumn{1}{c|}{0.498} & 0.295 & - \\ \hline
GSLoc, init with 5 desc.  & \multicolumn{1}{c|}{0.091} & 0.054 & \multicolumn{1}{c|}{\textbf{0.487}} & 0.286 & \textbf{35.52} \\ \hline
GSLoc, avg for all IoUs & \multicolumn{1}{c|}{\textbf{0.078}} & \textbf{0.035} & \multicolumn{1}{c|}{0.534} & \textbf{0.264} & 35.36 \\ \hline
\end{tabular}
}
\caption{Average pose estimation errors for successfully localized frames. GSLoc leads to comparable or even smaller pose errors compared with the sparse baseline method.}
\vspace{-6mm}
\label{tab:metrics}
\end{table}
\subsection{Real Data Results}
In the last experiment, we show that the results obtained for the synthetic data remain valid for the real-world datasets. To do so, we conducted similar experiments with camera initialization with NetVLAD descriptors and image base map rendering extension for 2 real indoor scenes from the Deep Blending dataset \cite{hedman2018deep}. Each scene is represented with a comprehensive set of arbitrary posed photos. 

\begin{figure}[thpb]
 \centering
 \vspace{1.75mm}
\includegraphics[width=0.88\linewidth]{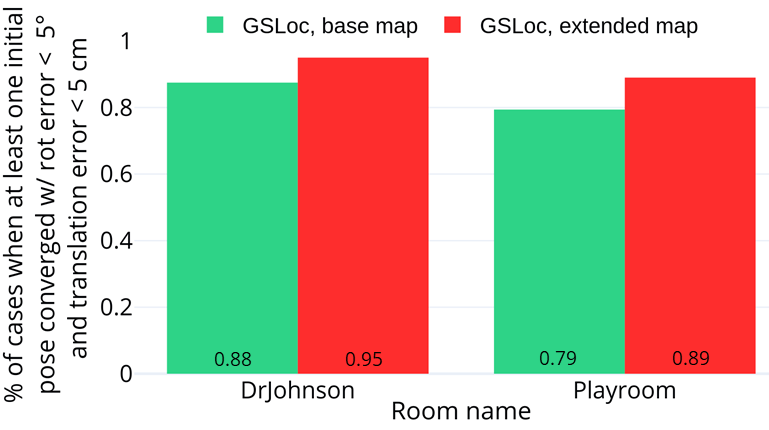}
\vspace{-1.5mm}
  \caption{Quantitive results on the real scenes from Deep Blending \cite{hedman2018deep} dataset. Enhancing the GSLoc camera initializations obtained by the image retrieval with the rendering-extended imagebase leads up to 10 $\%$ success rate improvement matching the observations obtained with synthetic data.}
  \vspace{-5mm}
  \label{fig:base_rebase_real}
\end{figure}

We form a test query set by taking each $8^{th}$ image of the set. We use the rest of the frames as the image base firstly for the SfM reconstruction and 3DGS training and secondly for the initial pose estimation with closest descriptors. Following the same experimental design, we again extend the image base with 3DGS renderings and showcase that the effectiveness of this technique remains valid also for the real scenes. We report our results in Fig.~\ref{fig:base_rebase_real}. 
We observe that the results achieved for the real scenes are consistent with those of the synthetic scenes. This is a strong indication of the  suitability of GSLoc for real-world visual localization.

%% file: 6_ablation.tex
\section{ABLATION STUDY}
\begin{figure}[ht]
 \centering
  \subfloat[Method ablation]{\includegraphics[width=.48\linewidth]{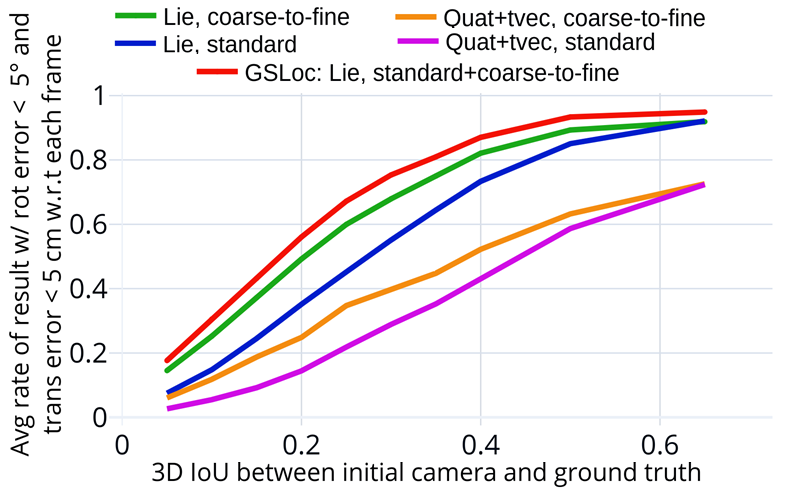}}
  \subfloat[Resolution ablation]{\includegraphics[width=.48\linewidth]{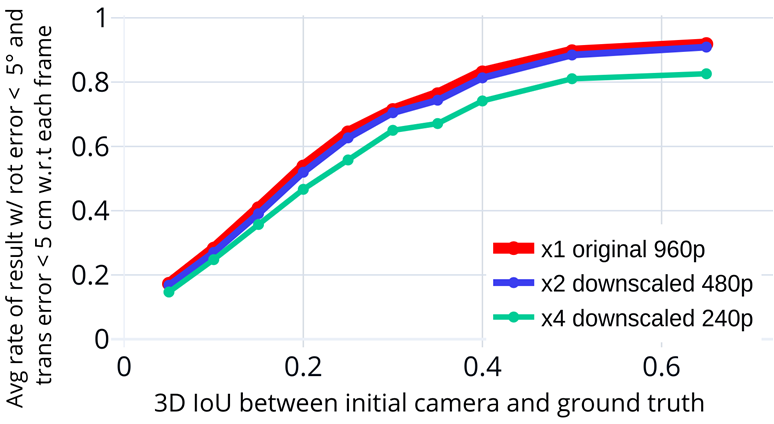}}
  \caption{Ablation study on optimization strategy and pose parametrization (a) and rendering resolution (b). Proposed two step standard+coarse-to-fine GSLoc optimization on manifold outperforms other methods and allows running on 2x downscaled images without results degradation.}
  \vspace{-4mm}
\label{fig:ablation}
\end{figure}

We perform a ablation of our GSLoc method by re-running the experiment described in section V.A utilizing different optimization strategies and pose parametrizations. We summarize the results in  Fig.~\ref{fig:ablation}(a) showing the advantage of the proposed two step standard-coarse-to-fine GSLoc on manifold optimization over other methods.

Furthermore, we estimate how the rendering resolution affects the GSLoc localization results, describing the details in Fig.~\ref{fig:ablation}(b). While we run all our experiments on an image resolution of 960x540 pixels we see that the decrease of image size results in almost identical results for 2x downscaling and starts suffering localization degradation only at a 4x image downscaling.

%% file: 7_limiatations_conclusions.tex
\section{LIMITATIONS AND FUTURE WORK}
While GSLoc shows promising results for visual localization, there are a few issues and limitations that we did not address in this work. An important one is the running-time efficiency of the method, which has not been optimized. Both, the multi-step sequential optimization as well as the use of a first order gradient descent method can lead to an increased execution time. In the future, we plan to improve the time efficiency of GSLoc and mitigate the current limitations by utilizing second order optimization algorithms.

\section{CONCLUSIONS}
We have presented GSLoc - a novel visual localization technique based on 3D Gaussian Splatting environment map representation. We have demonstrated both on synthetic and real data that our method is capable of performing accurate camera pose estimation. We have confirmed it via a comprehensive convergence analysis of various camera initializations and parametrizations. We have thoroughly explored the convergence limitations due to non-convexity of the photometric loss and proposed a coarse-to-fine strategy to mitigate this issue. Finally, we have proposed an effective way to improve the localization results by enhancing the GSLoc camera initialization, which is obtained by image retrieval with a refined image base that is extended with 3DGS-rendered camera frames.

%% file: 8_acknowledgements.tex
\section*{\small{ACKNOWLEDGEMENTS}}
\small{ The work was supported by the Analytical center under the RF Government (subsidy agreement 000000D730321P5Q0002, Grant No. 70-2021-00145 02.11.2021}